\documentclass{article}
\usepackage{PRIMEarxiv}

\usepackage[numbers]{natbib}
\usepackage{type1cm}        
\usepackage{makeidx}         
\usepackage{graphicx}        
\usepackage{multicol}        
\usepackage[bottom]{footmisc}
\usepackage{todonotes}
\usepackage{subcaption}
\usepackage{siunitx}
\usepackage{newtxtext}       %
\usepackage[varvw]{newtxmath}       
\usepackage{placeins}
\usepackage{lineno}

\makeindex             

\usepackage[utf8]{inputenc} 
\usepackage[T1]{fontenc}    
\usepackage{hyperref}       
\usepackage{url}            
\usepackage{booktabs}       
\usepackage{amsfonts}       
\usepackage{nicefrac}       
\usepackage{nicematrix}
\usepackage{microtype}      
\usepackage{lipsum}
\usepackage{fancyhdr}       
\usepackage{graphicx}       
\usepackage[linewidth=1pt]{mdframed}
\graphicspath{{media/}}     

\title{Deep Learning based Forecasting: a case study from the online fashion industry}

\author{
  Manuel Kunz \\
  Zalando SE \\
  \texttt{manuel.kunz@zalando.de} \\
   \And
  Stefan Birr \\
  Zalando SE \\
  \texttt{stefan.birr@zalando.de} \\
  \And
  Mones Raslan \\
  Zalando SE \\
  \texttt{mones.raslan@zaland.de} \\
  \And
  Lei Ma \\
  Zalando SE \\
  \texttt{lei.ma@zalando.de} \\
  \And
  Zhen Li \\
  Zalando SE \\
  \texttt{peter.zhen@zalando.de} \\
  \And
  Adele Gouttes \\
  Zalando SE \\
  \texttt{adele.gouttes@zalando.de} \\
  \And
  Mateusz Koren \\
  Zalando SE \\
  \texttt{mateusz.koren@zalando.de} \\
  \And
  Tofigh Naghibi \\
  Zalando SE \\
  \texttt{tofigh.naghibi@zalando.de} \\
  \And
  Johannes Stephan \\
  Zalando SE \\
  \texttt{johannes.stephan@zalando.de} \\
  \And
  Mariia Bulycheva \\
  Zalando SE \\
  \texttt{mariia.bulycheva@zalando.de} \\
  \And
  Matthias Grzeschik \\
  Zalando SE \\
  \texttt{matthias@zalando.de} \\
  \And
  Armin Kekić \thanks{Work done while at Zalando}\\
  Max Planck Institute for Intelligent Systems \\
  Tübingen, Germany \\
  \texttt{armin.kekic@mailbox.org} \\
  \And
  Michael Narodovitch \\
  Zalando SE \\
  \texttt{michael.narodovitch@zalando.de} \\
  \And
  Kashif Rasul \footnotemark[1]\\
  Morgan Stanley \\
  \texttt{kashif.rasul@gmail.com} \\
  \And
  Julian Sieber \\
  Zalando SE \\
  \texttt{julian.sieber@zalando.ie} \\
  \And
  Tim Januschowski \\
  Zalando SE \\
  \texttt{tim.januschowski@zalando.de} \\
}

\begin{document}
\maketitle

\begin{abstract}Demand forecasting in the online fashion industry is particularly amendable to global, data-driven forecasting models because of the industry's set of particular challenges. These include the volume of data, the irregularity, the high amount of turn-over in the catalog and the fixed inventory assumption. While standard deep learning forecasting approaches cater for many of these, the fixed inventory assumption requires a special treatment via controlling the relationship between price and demand closely.
In this case study, we describe the data and our modelling approach for this forecasting problem in detail and present empirical results that highlight the effectiveness of our approach.
\end{abstract}

\section{Introduction}
Forecasting problems in the fashion industry and in particular, in its online variant, come with a particular set of challenges and down-stream use-cases which we illustrate in our article. The arguably most peculiar challenge in the fashion industry stems from a fixed inventory assumption where a meaningful part of the inventory arrives at the beginning of a season (having been ordered much ahead of the season) and it cannot be re-ordered throughtout the season. Hence, for this non-reorderable part of the assortment, forecasting is not primarily used for replenishment use-cases as is the case in say, grocery (e.g.,~\citep{FILDES20221283}). 
Instead, a primary use-case for forecasting is pricing/discounting (e.g.,~\citep{Falcon20}). 
Forecasting provides the input for a down-stream mixed integer optimization problem that sets the prices with a notion of optimality in a classical forecast-then-optimize setting. In the online fashion industry, solving the pricing problem is a means to handle inventory risk management because pricing presents a major lever to reduce the risk to be overstocked at the end of the season. 
Complications that arise from the fixed inventory assumptions include the prominence of cold start problems (a large part of the catalogue is renewed each season) and short history. Other complications are commonalities also present in other online industries, namely a large catalogue with millions of stock keeping units (SKUs) which on the one hand results in a large overall scale of the forecasting problem and on the other hand in sparsity on the SKU level. We present examples and illustrations of the complications in a detailed (necessarily anecdotal) description of the data and the implied challenges for forecasting.

The main contribution of our article is the description of the approach that Zalando SE, a market-leading fashion retailer in Europe, has taken to address the forecasting problems that arise in pricing. Demand forecast subject to differing pricing levels poses the primary forecasting challenge that we aim to address. Given the complications described above, a natural choice is a flexible and hence, highly-parameterized, global~\citep{JANUSCHOWSKI2020167} forecasting model that allows to learn complex patterns across products and seasons in a data-driven way. We describe a formerly unpublished deep learning based forecasting model relying on the transformer~\citep{vaswani2017attention} architecture. It has been in use at Zalando in different versions since 2019 and hence is, to the best of our knowledge, among the first examples of attention-based forecasting models in production in industry. 

Our article is structured as follows. We start by describing the data and associated covariates in Section~\ref{sec:data}. In Section~\ref{sec:model}, we describe our model in detail and explain how we address the problems that arise in our context.  We place a particular focus on the importance of the rich covariate structure available in our forecasting problems. The importance of price as a covariate is fundamental and we describe how we incorporate inductive biases to learn appropriate  price elasticities into our model. In Section~\ref{sec:empirical_result}, we include empirical results and compare our model performance against other approaches. We also try to explain why a transformer-based architecture performs better for our use-case, by testing for scaling laws. We discuss related work in Section~\ref{sec:rel_work}, where we mention advantages and limitations of our approach, contrasting our approach with publicly available methods. In Section~\ref{sec:practice}, we comment on practical challenges such as monitoring forecasting accuracy in production, its down-stream usage and deciding how to re-train. We conclude in Section~\ref{sec:concl}, where we also discuss avenues for future work that hopefully inspire future research.

\section{Data for forecasting at Zalando: An Overview}\label{sec:data}
\begin{figure}
    \includegraphics[width=\textwidth]{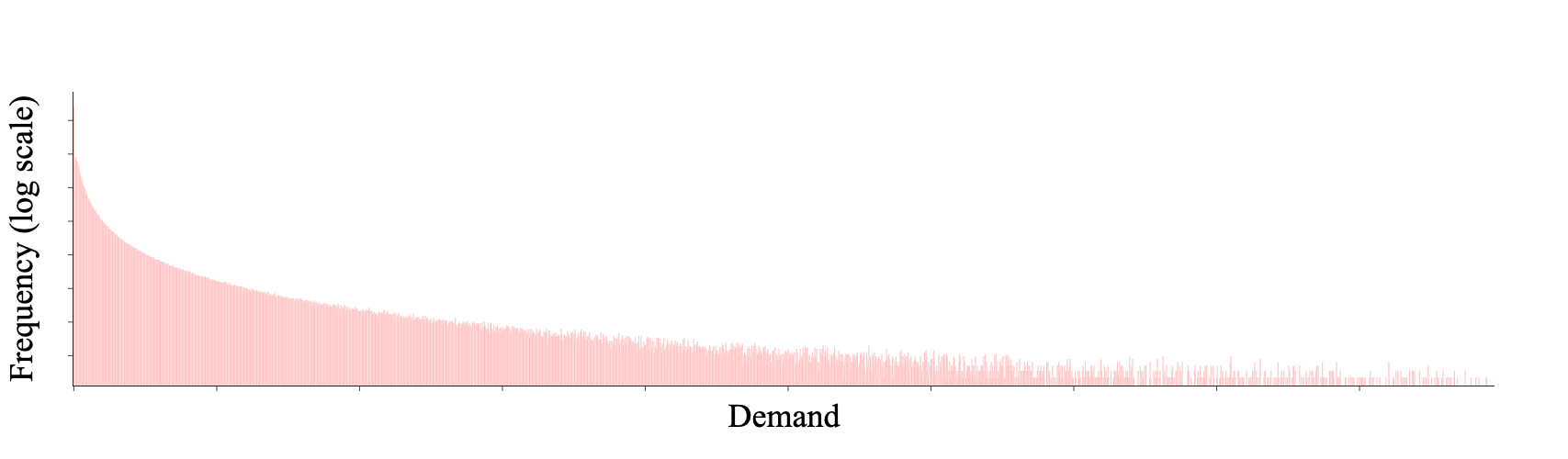}
    \caption{Histogram of demand (in log scale) per article: The histogram shows the heavy tailed demand distribution. On the x-axis is demand starting from 0, on the y-axis the frequency of the demand occurrence over one time unit.}
    \label{fig:demand_hist}
\end{figure}

Zalando is active in 25 European markets, its catalog comprises of >6500 brands with >50M active customers. This amounts to 1.6 million articles on the Zalando platform (as of Q3 2021). We are interested primarily in modelling the future demand of an article, where the demand is the quantity of items customers would buy in one time unit. 
Demand at Zalando follows a typical long-tail distribution, see Figure~\ref{fig:demand_hist}. Note that we cropped the histogram on the right and cannot provide units for the x-axis and y-axis for confidentiality reasons. Most articles do not sell on most days, but there are a few high-selling products. Similar demand characteristics are available in other real-world examples~\citep{salinas2020deepar,lim2021temporal,laptev2017}.

\begin{figure}
\centering
\begin{subfigure}[b]{\textwidth}
    \includegraphics[width=\textwidth]{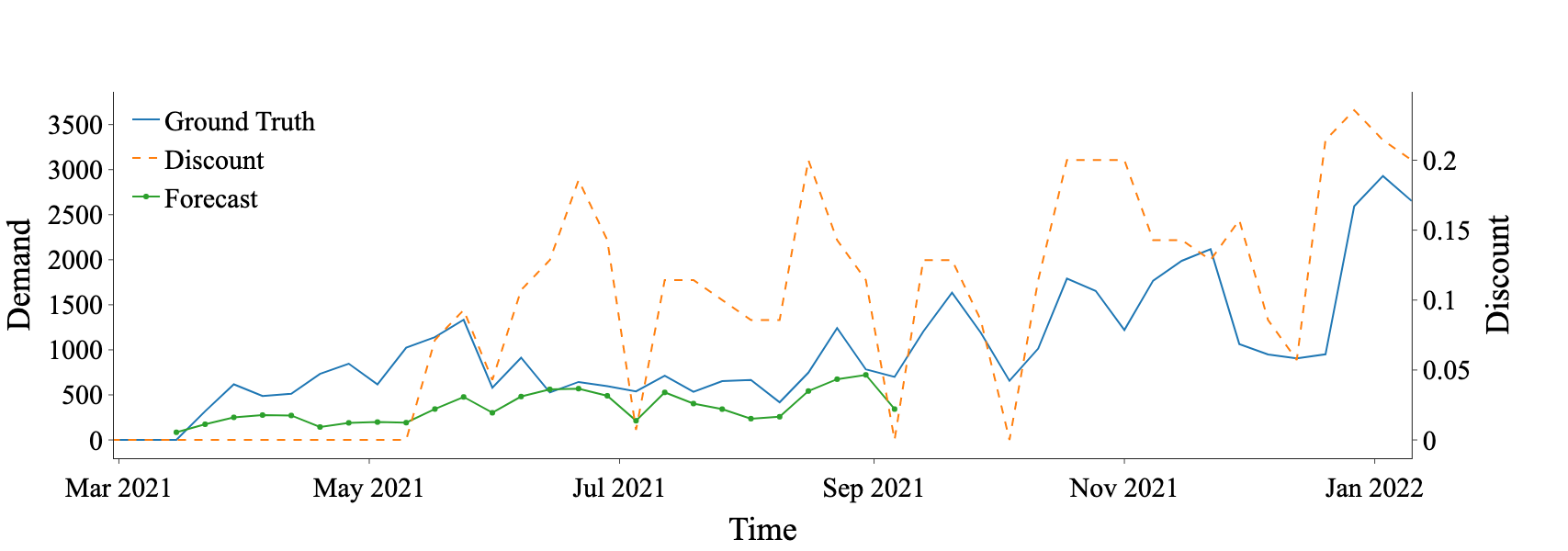}
    \caption{Example of a cold start problem.}
    \label{fig:cold_start}
\end{subfigure}
\begin{subfigure}[b]{\textwidth}
    \includegraphics[width=\textwidth]{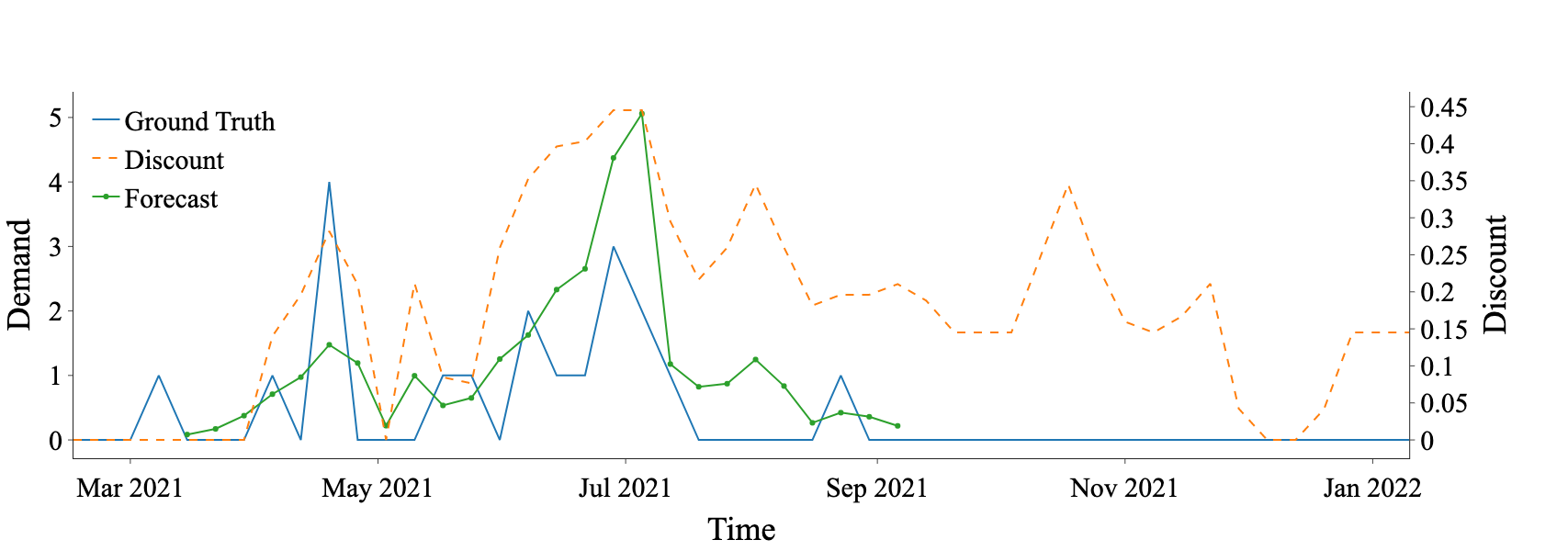}
    \caption{Example of short history.}
    \label{fig:short_hist}
\end{subfigure}
\begin{subfigure}[b]{\textwidth}
    \includegraphics[width=\textwidth]{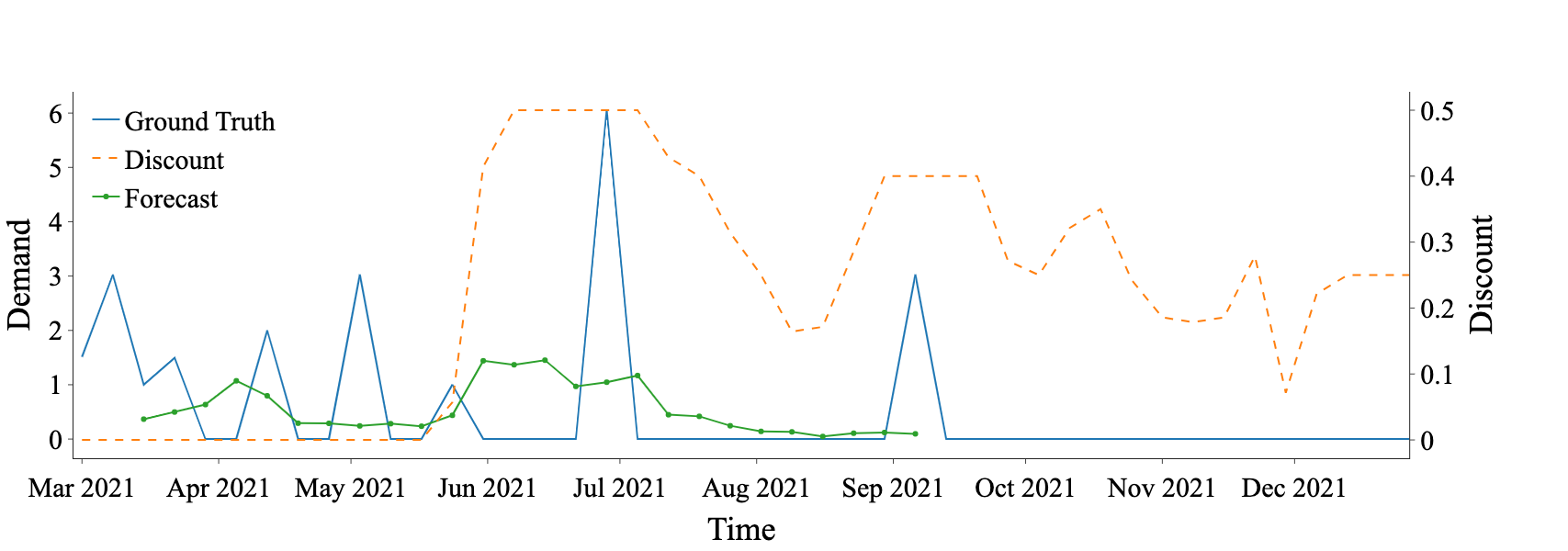}
    \caption{Example of a sparsity/intermittency.}
    \label{fig:sparse}
\end{subfigure}
\caption{Examples for complications occurring in practice: The dashed orange line gives discount level (the higher, the more discount), blue is observed (ground truth) demand and green with markers is the forecast. Note that everything left to the first green dot is the available demand history. We have hence articles with little to no history, see Figures~\ref{fig:cold_start} and~\ref{fig:short_hist} as well as sparse time series, Figure~\ref{fig:sparse}.}
\label{fig:cSKU_examples}
\end{figure}

\begin{figure}
    \centering
    \includegraphics[width=\textwidth]{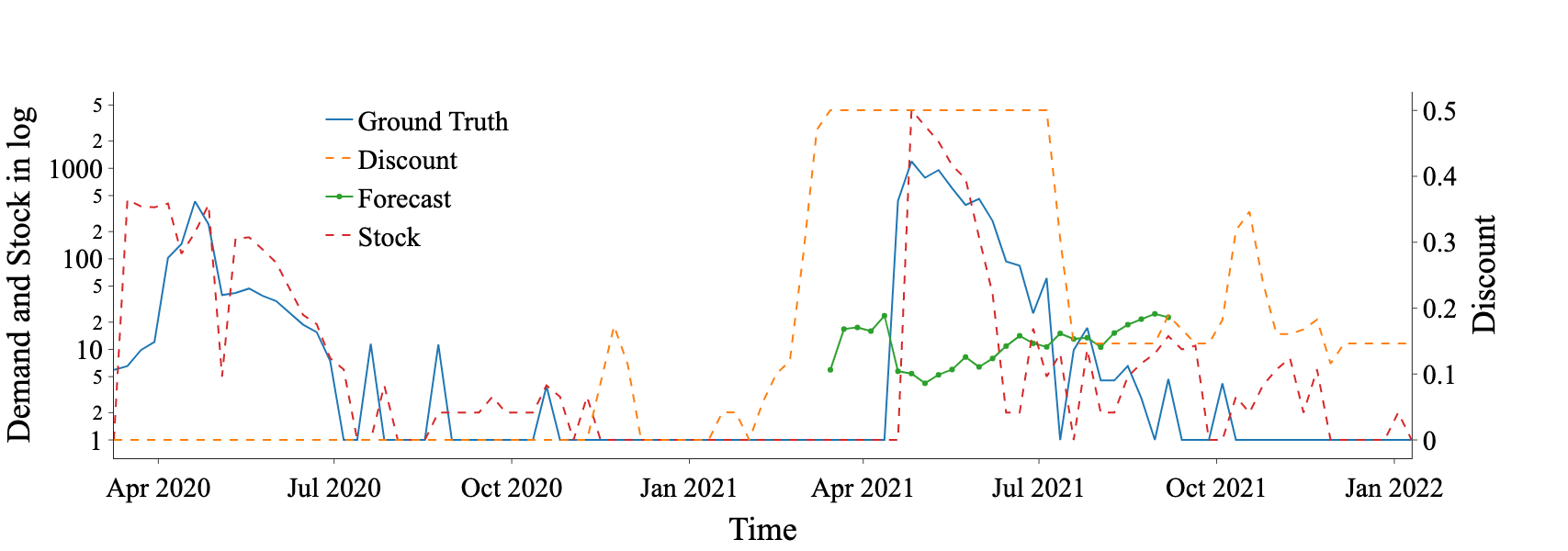}
    \caption{Example for a challenging stock situation: The red dashed line represents stock levels over time. The article was out of stock for a longer time period until it was replenished in April 2021. It can be seen that the ground truth is highly correlated with stock availability of an article which provides additional modelling challenges as discussed in Section~\ref{sec:pecularities}}
    \label{fig:oos}
\end{figure}

Complementing the aggregate statistics, Figure~\ref{fig:cSKU_examples} gives examples for what individual time series look like. In Figure~\ref{fig:oos}, we show additionally the stock levels in red dashed lines for an example of an article that has a complicated stock availability. These examples illustrate the difficulties in extracting patterns from the observational time horizon of a single article.


\subsection{Sales to Demand Translation}\label{sec:pecularities}
Note that, similar to other demand forecasting problems (e.g.,~\citep{FILDES20221283,wupper17,MQRNN,januschowski19}), we do not observe demand, but only sales. Demand corresponds to what the customers want to purchase, and materialises only when enough stock is available. In our case, we have a stock signal available which indicates when sales are strictly less than demand because no stock is available. While it is possible to handle such cases by marking the demand data as \emph{missing} and let the forecast model handle this case (see~\citep{resilient} for an example), here we adopt a different approach by pre-processing our sales data and impute demand.

\begin{figure}
    \centering
    \includegraphics[width=\textwidth]{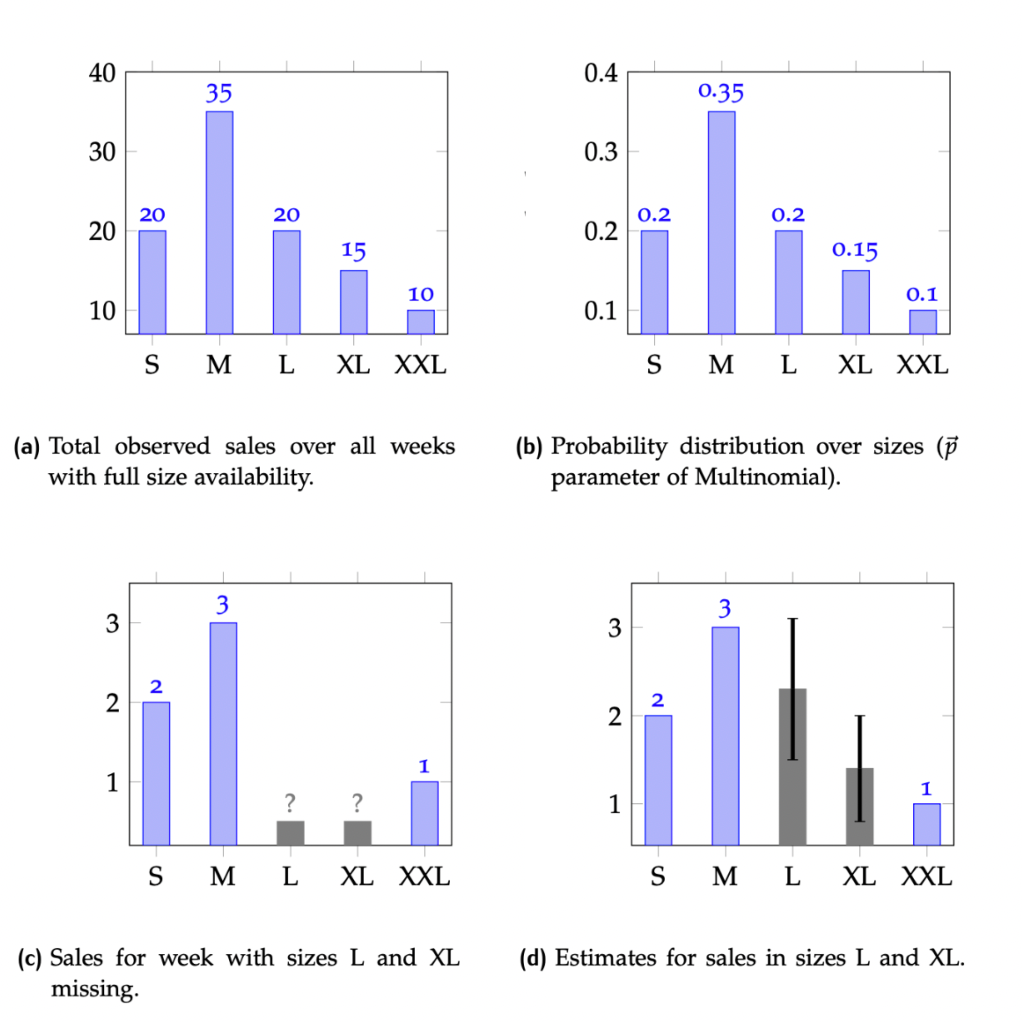}
    \caption{Sales to demand translation for an article with 5 sizes: In (a), historical sales observations of an article over weeks with all sizes available are aggregated. Based on (a), we obtain in (b) the articles's empirical probability distribution over sizes. In (c), the weekly demand of an article with missing sizes L and XL is illustrated. The unobserved demand for L and XL is inferred in (d), given the observed demand of the other sizes for that week and the empirical distribution computed in (b).}
    \label{fig:sales2demand}
\end{figure}

\begin{figure}
    \centering
    \includegraphics[width=\textwidth]{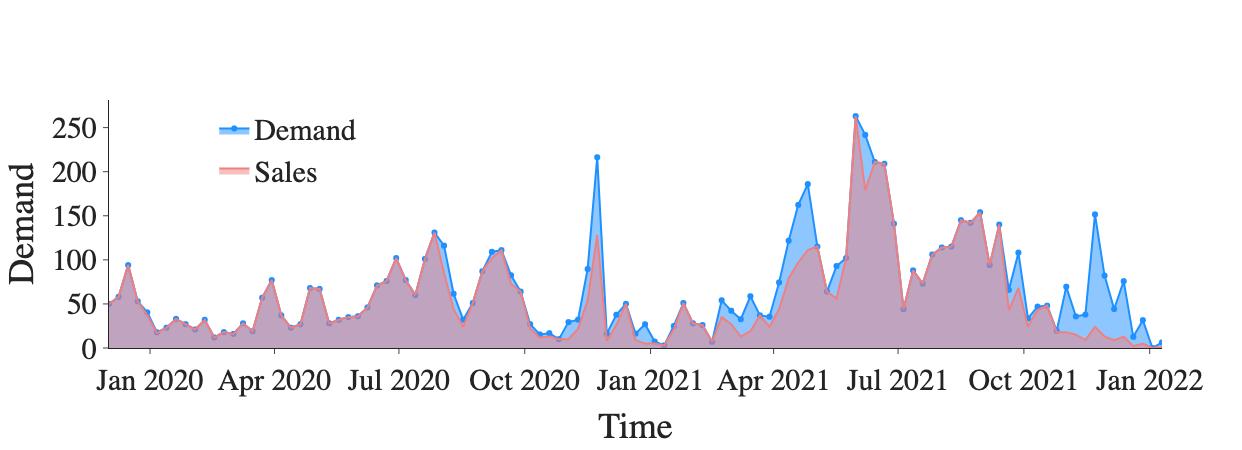}
    \caption{Demand vs sales for a single article. In weeks with full availability, sales and demand overlaps. If a size or the full article becomes unavailable, demand becomes an estimate larger than the observed sales.
    }
    \label{fig:demand_vs_sales}
\end{figure}

We can take advantage on the fact that in fashion, articles typically come in different sizes. Note that we do not desire to price differently different sizes, and so the forecasting task requires to predict demand of a single fashion item aggregated over all sizes. If all sizes of an article are available or \emph{in stock} over one time unit, the demand of that article equals  the materialized/observed sales. If some sizes are out of stock, the demand is in fact (partially) unobserved and we cannot directly obtain it from the sales data. To remediate this, we first define the demand $q$ of an article $i$ with $k$ sizes as a random variable 
$\bold{q_{i}} = \{q_{i1}, q_{i2}, \ldots, q_{ik}\}$. In order to infer the expected demand over all sizes 
$n_i = \sum_{n=1}^k q_{in}$, we assume $\bold{q_{i}}$ to be multinomially distributed with 
$\bold{q_{i}} \sim \mathrm{Multinomial}(n_i, \bold{p_i})$ and $\bold{p_i} = \{p_1, p_2, \ldots, p_k\}$ 
the discrete probability distribution over the $k$ sizes of article $i$. If we further assume to know $\bold{p_i}$, we can compute $n_i$ based on a partial observation 
$\{q_{i1}, q_{i2}, \ldots, X_{ij},\ldots, q_{ik}\}$, with $X_{ij}$ representing the missing demand observations. 
This procedure only requires to learn $\bold{p_i}$ from observational data in intervals with no stock-outs, assuming $\bold{p_i}$ to be time invariant. 
We then use the learned $\bold{p_i}$ to translate sales to demand, and we use this imputed demand in model training.
Note that, when adopting such a preprocessing step, special care has to be taken by not evaluating our model on extrapolated data, but rather exclude this data from evaluation. In Figure~\ref{fig:sales2demand}, we depict the process described above and Figure~\ref{fig:demand_vs_sales} visualizes the result of our model applied to a specific article over time.

\subsection{Description of Covariates}
The covariates that we use in our forecasting model can be grouped into the following categories: \emph{static-global}: time-independent \& market-independent; \emph{dynamic-global}: time-dependent \& market-independent; \emph{dynamic-international}: time-dependent \& market-specific; and \emph{static-international}: time-independent \& market-specific. We give a non-exhaustive list for the covariates in use, described in Table~\ref{tab:covariates} in more detail for illustration purposes.
\begin{table*}[t]
	\begin{center}
		\begin{scriptsize}
			\begin{sc}
				\begin{tabular}{lccccc}
    \toprule
Feature	& Dim &	Category &	Value Type &	Transformation & available in future? \\
\midrule
Brand	& 10	& static-global & 	categorical	& embedded & y \\
Commodity Group &	10	& static-global &	categorical &	embedded & y \\
Discount &	14 &	dynamic-int’l &	numeric &	– & n \\
Black Price & 	14	& static-int’l &	numeric	& logarithm & y \\
Sales &	14	& dynamic-int’l &	numeric &	logarithm & n \\
Stock &	1 &	dynamic-global &	numeric &	logarithm & n \\
Stock Uplift & 	1 &	dynamic-global &	numeric	& logarithm & n \\
                \bottomrule
                    \end{tabular}
                \end{sc}

                \caption{Examples for covariates available for the forecasting problem at Zalando.}
                \label{tab:covariates}
            \end{scriptsize}
    \end{center}
\end{table*}

The \emph{static-global} covariates Brand and Commodity Group categorize an article. The later provides information about the part of the assortment (e.g, Men - Sportswear). They do not change over time and across countries. The Black Price is the recommended retail price and the initial price an article receives. It is the reference price for every discount applied. We use its logarithm to prevent scaling issues typical in global forecasting models~\citep{JANUSCHOWSKI2020167}. Black Price and Discount are \emph{international} covariates, containing potentially different values in different countries. The Black Price of an article does not change over time, but Discount might change which makes it \emph{dynamic}. 

Note that not all covariates are available for the forecast horizon. For example, Stock or Discount are only available historically. We include them because they either help to "explain away" certain historical effects or because we set them explicitly in the forecast horizon to obtain what-if forecasts. Other covariates, like the Stock Uplift, are only available as an estimate for the future time horizon. The Stock Uplift expresses for each time unit the increase in stock of an article, caused by supplier deliveries or customer returning their pruchase. This value is available for the past but only partially available for the future and needs to be estimated based on supplier delivery data. Both Stock and Stock Uplift are \emph{global}, because stock is shared across countries.

\section{Demand Forecast Model}\label{sec:model}
The demand forecasting model used within Zalando Pricing is a global~\citep{MONTEROMANSO20211632,JANUSCHOWSKI2020167} forecasting model. This means that a single forecasting models is trained over the entire  Zalando article\footnote{A sellable article offered in the Zalando fashion store.} assortment and used to provide article specific predictions on a weekly time resolution for a horizon of 26 weeks. Additionally, our forecast is used to decide on the discounts applied to each article: we need to forecast future demand for multiple scenarios. Before we describe the forecasting approach in detail, we first formalize the forecasting problem.

\subsection{Problem Formalization}
For any timeseries $x$, $x_{0:T}$ is short-hand for $[x_0, x_1, \ldots, x_T]$. 
The observational timeseries data of an article $i$ at time $t$ starting at $0$ is given by $\{q_{i0:t}, d_{i0:t}, z_{i0:t}\}$, where 
$q$ denotes the demand. $d$ corresponds to the discount, which is the percentage of price reduction relative to the article's recommended retailer price; and $z$ a set of article specific covariates. 
Our object of interest is 
\begin{equation}\label{eq:demand_model_prob}
P(q_{i,t+1:t+h}| q_{i,0:t}, d_{i,0:t+h}, z_{i,0:t+h}; \theta)\;,
\end{equation}
that is the probability distribution of demand in the forecast horizon $t+1:t+h$ conditioned on (among others) discounts in the forecast horizon. We are interested in identifying appropriate $\theta$. Note that~\eqref{eq:demand_model_prob} is a simplification. More generally, we should be interested on the one hand in a multi-variate version so that we model cross-article effects and, on the other hand, we should actually be interested in

\begin{equation}\label{eq:demand_model_prob_causal}
P(q_{i,t+1:t+h}| \mathrm{do} (d_{t+1:t+h}), q_{0:t}, d_{0:t}, z_{0:t+h}; \theta)\;,
\end{equation}

where the $\mathrm{do}$ operator denotes the intervention of setting a price/discount to a certain level (see e.g.,~\citep{reason:Pearl09a} for the notation and additional background). This is because we are interested in taking down-stream decisions on prices so the (causal) relationship between price/discount and demand is of fundamental importance. 
We leave a more comprehensive causal treatment for future work and mention that in practice, we approximate~\eqref{eq:demand_model_prob} by a point forecast and point to~\citep{adele21} for a discussion of a probabilistic extension of the model described here. However, we remark that~\eqref{eq:demand_model_prob} is a simplification of our actual model in the sense that we provide a 
multi-variate forecast because we forecast all markets in which Zalando is active in simultaneously. So our demand time series actually has an additional index per market. However, for ease of exposition, we do not further elaborate on this complication.

\subsection{Model Architecture}
For the demand forecast model we rely on an encode/decoder~\cite{sutskever2014sequence} approach. This is primarily motivated by the fact that we have some covariates that are not available in the forecast horizon and others that are available only for the future, see Table~\ref{tab:covariates}. The encoder/decoder architecture is particularly amendable for this setting, see e.g.,~\citep{januschowski19,MQRNN}. 

\begin{figure}
    \centering
    \includegraphics[width=\textwidth]{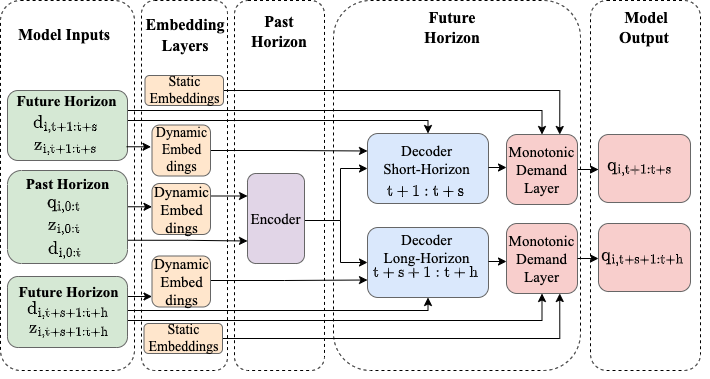}
    \caption{Demand Forecaster Architecture: Encoder handles the observational time series, decoder incorporates the future covariates. Future discounts are directly provided to the output (Monotonic Demand) layer, skipping decoder and encoder.}
    \label{fig:demand_fc_arc}
\end{figure}

For the model architecture itself, we base it on the standard Transformer architecture and we depict it in Figure~\ref{fig:demand_fc_arc}. We have additional embedding~\cite{mikolov2013efficient} layers for some of the covariates. 
Our main modelling innovations are two-fold. First, we specialize our decoder into a short- and a long-term horizon. This is because our down-stream application requires particular focus on the first forecast weeks for which the pricing decisions are taken, whereas the further out future only plays a secondary importance. We further adapt the training scheme to account for the focus on the first weeks. We discuss this in Section~\ref{sec:training}. Second, we need to control the relationship between price and demand closely given the primary usage of the forecasts in pricing. We achieve this by enforcing a 
monotonic relationship between price and demand which we parameterize flexibly as a piece-wise linear function (Section~\ref{monotoniclayer}).

\subsubsection{Input preparation}\label{sec:input_preparation}
In the forward pass, the model receives the data of a single article shown as Model Inputs in Figure~\ref{fig:demand_fc_arc}. All features are stacked into a single input tensor of dimension $\mathbb{R}^{t\times u \times v}$, where $t$ is the time dimension, $u$ the batch size and $v$ the dimension of the covariate vector.

The high dimensional but sparse categorical covariates are passed through the embedding layers before being forwarded to the encoder, decoder and monotonic demand layer. We distinguish between two types of embeddings, dynamic and static. Dynamic embeddings are time dependent and provided as input to the encoder and decoder. Static embeddings are time independent and provided as additional input to the monotonic demand layer. The temporal structure of the time series is represented by positional encoding, introduced in Section~\ref{sec:pos_enc}.

The decoder receives all covariates associated with the forecasting time horizon, the associated embeddings and the output of the encoder. 
Encoder and decoder consist of various multihead attention layers~\citep{bahdanau2014neural, vaswani2017attention}, to represent temporal correlations across the covariates. Instead of feeding future discounts into the decoder, we directly provide them to the monotonic demand layer, see Section~\ref{monotoniclayer}, which allows us to predict demand as a monotonically increasing function of discount.

\subsubsection{Encoder}\label{encoder_decoder}
The task of the encoder is to learn a representation of the observational article data, optimized for predicting future demand. The encoder model architecture is based on multi-head attention~\cite{vaswani2017attention}. In the following, we recall the concepts and explain how to translate them to the timeseries setting. In a first step, we project the input data $\{d_{0:t}, q_{0:t}, z_{0:t}\}$ into three $(Q, K, V)$ \footnote{$Q =$ Query, $K =$ Key, $V =$ Value as in the standard transformer nomenclature; note that this clashes with the notation that we introduced before unfortunately, but it is standard in the transformer literature.} tensors. Therefore, for each article $i$, the data is transformed into the matrix $\eta_{{0:t}}$: 

\begin{equation}\label{eq:enc_in}
\eta_{i,{0:t}} = 
\begin{bmatrix}
   q_{00} & q_{01} & \cdots & q_{0t} \\
   d_{10} & d_{11} & \cdots & d_{1t} \\
   z_{20} & z_{21} & \cdots & z_{2t}\\
   \vdots  & \vdots  & \ddots & \vdots  \\
   z_{v0} & z_{v1} & \cdots & z_{vt} 
\end{bmatrix}, Q = \phi_q(\eta_{i,0:t}), K = \phi_k(\eta_{i,{0:t}}), V = \phi_v(\eta_{i,{0:t}})
\end{equation}

Each column in $\eta_{i,0:t}$ represents a covariate vector with demand, discount and additional covariates of an article $i$ at time $T \in [0, \ldots, t]$.
The matrix $\eta_{i,0:t}$ is then multiplied by three learned weight matrices, implemented as linear layers $\phi_q, \phi_k, \phi_v$.
Scaled dot product attention~\eqref{eq:self_attention} and multi-head attention~\eqref{eq:multi_head_attetion} is then applied to $Q, K$ and $V$, following~\citep{vaswani2017attention}.

\begin{equation}\label{eq:self_attention}
\mathrm{Attention}(Q, K, V) = \mathrm{softmax}(\frac{QK^T}{\sqrt{d_k}})V
\end{equation}

\begin{equation}\label{eq:multi_head_attetion}
\begin{aligned}
\mathrm{MultiHead}(Q, K, V) &= \mathrm{Concat(head_1, ..., head_h})W^O \\
\mathrm{head}_j &= \mathrm{Attention}(QW_j^Q, KW_j^K, VW_j^V)
\end{aligned}
\end{equation}

In the original transformer architecture, $d_{model}$ represents the embedding dimension of an embedded word vector. Applied to the time series data set in~\ref{eq:enc_in}, $d_{model}$ is defined by the number of rows in $\eta_{{i,0:t}}$, or expressed differently, by the covariate vector of dimension $v$, so $v = d_{model}$. Similar as in the original transformer architecture, the number of attention heads $h$, the dimension of the linear layers $d_k$ and the dimension of the embedding layers were chosen to match $d_{model} = h \cdot d_k$. The linear layers $\phi_q, \phi_k, \phi_v$ were specified with same input and output dimension to produce $Q, K$ and $V$ tensors of shape $h \times (t+1) \times d_k$. 
The remaining operations follow~\eqref{eq:self_attention},\eqref{eq:multi_head_attetion}. Note that the computation of the $h$ attention heads allows parallelization and is most efficiently implemented by applying vectorized tensor operations~\cite{paszke2019pytorch}. The encoder contains multiple stacked multi-head attention layers with two sublayers and residual connections~\cite{he2016deep} as specified in~\cite{vaswani2017attention}. Dropout~\cite{srivastava2014dropout} and Layer Normalization~\citep{ba2016layer} is applied on the output of each sublayer.

\subsubsection{Positional Encoding}\label{sec:pos_enc}
Positional encoding is a fundamental part of the original transformer architecture. For each position in the input embeddings, a positional encoding of dimension $d_{model}$ is computed and summed up with the corresponding word embedding at this position~\cite{vaswani2017attention}. This approach is not transferable to the time series data introduced in Section~\ref{sec:data} because of the heterogeneous covariate structure. Instead, a positional encoding feature with embedding dimension $e_{dim}$ is computed and integrated as an independent covariate to $\eta_{i,{0:t}}$.

\begin{equation}\label{eq:pos_inc}
\begin{aligned}
    \mathrm{pos}_{0:t} &=
    \begin{bmatrix}
        \mathrm{pos}_{0}^T & \mathrm{pos}_{1}^T & \cdots & \mathrm{pos}_{n}^T & \cdots & \mathrm{pos}_{t}^T\\
    \end{bmatrix} \\
    \mathrm{pos}_n &= [\sin(p_{n,0}), \cos(p_{n,1}), \ldots, \sin(p_{n, e_{\dim}-2}), \cos(p_{n, e_{\dim}-1})] \\
    p_{n,m} &= 2 \cdot \pi \cdot f_m \cdot n \\
    f_m &= \frac{2 \cdot m + 1}{e_{\dim} \cdot t_{\dim}} 
\end{aligned}
\end{equation}

The positional encoding feature is defined by $pos_{0:t}$ in~\eqref{eq:pos_inc}. Each position in $\mathrm{pos}_{0:t}$ is a sequence of sine and cosine values computed at different frequencies $f_m$ for each embedding dimension in the positional encoding. Each frequency is defined relative to $t_{dim} = 52$, representing annual cycles given weekly time resolution.

\subsubsection{Padding and Masking}
The original transformer architecture applies zero padding and masking to sequences with variable length in order to reach the defined input embedding dimension and to avoid a look-ahead in the decoder. In time series forecasting, only the time dynamic covariates need to be zero padded and masked which becomes especially important if slicing is applied in order generate multiple training samples out of one observational time series. Masking is then applied to the result of the dot product of the $Q$ and $V$ tensors in order to prevent the padded dimensions influencing the attention scores. In the case of the demand forecasting model, not only zero padded values are masked out but also all data points with zero stock availability. This is motivated by issues inferring demand on article data with zero stock availability.

\subsubsection{Decoder}
The task of the decoder is to compute for each time unit in the forecast horizon a future state. The matrix $\eta_{i, t+1:t+h}$~\ref{eq:dec_in_2} represents the input to the decoder. Each column vector $\eta_{i,T}$ in $\eta_{i, t+1:t+h}$ contains the known future covariates of an article $i$ in future week $T \in [t+1, \ldots, t+h]$ and the last observed discount $d_t$ and demand $q_t$ at time $t$. In order to compute the future state at time $T$, the decoder receives the encoder's output $\gamma_{i,0:t}$ and the future covariate vector $\eta_{i,T}$~\eqref{eq:dec_in} as input. Similar to the encoder, the decoder contain multi-head attention layers but with a specific attention mechanism~\eqref{eq:source_attention} applied.

\begin{equation}\label{eq:dec_in_2}
\gamma_{i,0:t} = 
\begin{bmatrix}
   e_{00}& e_{01} & \cdots & e_{0t} \\
   e_{10}& e_{11} & \cdots & e_{1t} \\
   e_{20}& e_{21} & \cdots & e_{2t} \\
   \vdots  & \vdots  & \ddots & \vdots  \\
   e_{k0} & e_{k1} & \cdots & e_{kt} 
\end{bmatrix}, \eta_{i, t+1:t+h} = 
\begin{bmatrix}
   d_{0t} &  d_{0t} & \cdots &  d_{0t} \\
   q_{1t}&  q_{1t} & \cdots &  q_{1t} \\
   z_{2,t+1} &  z_{2,t+2}  & \cdots &  z_{2,t+h} \\
   \vdots & \vdots &  \ddots & \vdots \\
   z_{k,t+1} &  z_{k,t+2}  & \cdots & z_{k,t+h} \\
\end{bmatrix}
\end{equation}

\begin{equation}\label{eq:dec_in}
Q = \phi_q(\eta_{i, T}), \\ K = \phi_k(\gamma_{i,0:t}), V = \phi_v(\gamma_{i,0:t})
\end{equation}

In~\eqref{eq:dec_in}, the decoder input is passed through a linear layer $\phi$, similar as in~\eqref{eq:enc_in}. Note that the dimension of $\eta_{i,T}$ and $\gamma_{i,0:t}$ are different, causing $Q$ to be different in dimension compared to $K$ and $V$. This is solved by repeating values of $Q$ to align in dimension, before stacked together and passed through linear layer $\tau$ computing the inner product across past and future.

\begin{equation}\label{eq:source_attention}
\mathrm{Attention}(Q, K, V) = \mathrm{softmax}(\frac{\tau(Q, K)}{\sqrt{d_k}})V
\end{equation}

The slightly changed attention formula shown in~\eqref{eq:source_attention} is applied for each column vector in $\eta_{i, t+1:t+h}$, computing the future states in the forecast horizon $t+1:t+h$. Different to the original transformer architecture, the decoder is non-autoregressive, allowing to compute each future state in parallel. This architectural choice was necessary to enforce independence\footnote{The independence assumption between future weeks was made in order simplify the downstream pricing models, consuming the output of the demand forecaster.} between forecast weeks. Similar as in the encoder, normalization, dropout and residual connections are applied. The decoder output $\kappa_{i,t+1:t:h}$ is passed with the encoder state $\gamma_{i,0:t}$ and the future discounts $d_{i,t+1:t+h}$ to the Monotonic Demand Layer in order to predict the future demand $q_{i,t+1:t+h}$.

\subsubsection{Monotonic Demand Layer}\label{monotoniclayer}
The demand forecast model represents future demand as a function of future discount $d_{i,t+1:t+h}$. This requires not only to learn the temporally dependent dynamics of the demand time series but also the demand response function w.r.t. discount.

\begin{figure}
    \centering
    \includegraphics[width=\textwidth]{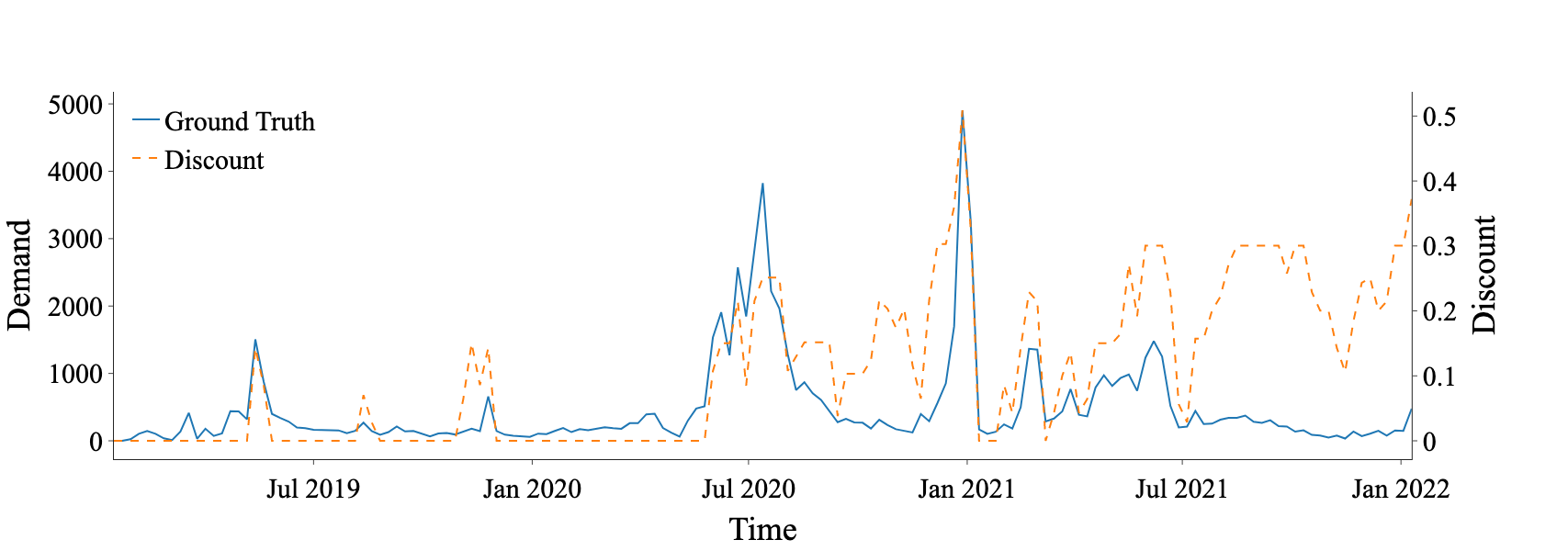}
    \caption{Demand and discount time series of an article, illustrating the positive correlation between discount (orange dashed) and demand (blue).}
    \label{fig:single_csku_time_series}
\end{figure}

Figure~\ref{fig:single_csku_time_series} shows demand and discount time series, illustrating the positive correlation between the two quantities. Although it might be obvious that the demand always increases with an increase in discount, the underlying causal mechanism is subject to confounding, resulting in counter intuitive demand and discount curves. In order to address this issue, we model demand $q_{i,t+n}$ of future discount $d_{i,t+n}$ as a piece-wise linear and monotonically increasing demand response function in the monotonic demand layer.

\begin{figure}
    \centering
    \includegraphics[width=8cm]{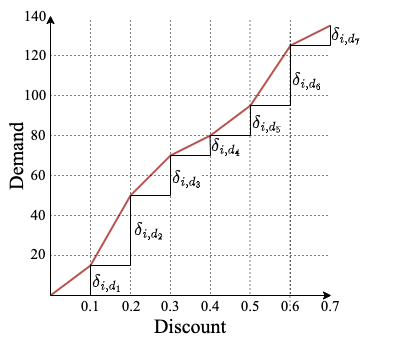}
    \caption{Piecewise linear monotonic demand response function as defined  in~\eqref{eq:monotonic_demand_response}, x-Axis: the discount level, y-axis: The increase in demand under the applied discount. The $\delta$ values represent the change in demand for a 10 percentage point change in discount.}
    \label{fig:montonic_demand_function}
\end{figure}

The domain of the discount dependent function shown in Figure~\ref{fig:montonic_demand_function} is divided into segments of equal width. Each segment contains a linear function, specifying the increase in demand relative to the black price demand. We assume the following parameterization of the demand response function $\xi$:

\begin{equation}\label{eq:monotonic_demand_response}
\begin{split}
\xi(d_{i,t+n}) = \hat{q}_{i,t+n} + \sigma_{i,t+n} \cdot \Big(\sum_{n=1}^{m}\delta_{i,d_n} + d_{i,t+n} \cdot \frac{\delta_{i,d_{m+1}}}{0.1}\Big) \\
 d_{i,t+n} \in [0,0.7], m = \lfloor 10 \cdot d_{i,t+n}\rfloor
\end{split}
\end{equation}

The monotonic demand layer parameterizes the demand function defined in~\eqref{eq:monotonic_demand_response} via two feed forward neural networks, $\phi_0$ and $\phi_1$

\begin{equation} \label{eq:monotonic_layer}
    \begin{bmatrix}
       \delta_{id_1} \\
       \delta_{id_2} \\
       \vdots \\
       \delta_{id_7} \\
    \end{bmatrix} = \phi_0(\gamma_{i,0:t}), \\
    \begin{bmatrix}
       \hat{q}_{i,t+n} \\
       \sigma_{i,t+n} \\
    \end{bmatrix} = \phi_1(\kappa_{i, t+n})\;. \\
\end{equation}

The input to $\phi_0$ and $\phi_1$ is the output of encoder and decoder, $\gamma_{i,0:t}$ and $\kappa_{i, t+n}$. 
The slopes $\delta_{id_1}, \ldots, \delta_{id_7}$ only depend on the encoder state, resulting in the same values for all future weeks. Black Price (see Table~\ref{tab:covariates}) demand $\hat{q}_{i,t+n}$ and scale parameter $\sigma_{i,t+n}$ are dependent on the decoder output column $\kappa_{i, t+n}$, indexed with $t+n$, similar as the future discount $d_{i,t+n}$. The Softplus\footnote{Smoothed out version of the ReLU~\citep{nair2010rectified} function.} function ensures that the output of $\phi_0$ and $\phi_1$ results in a monotonically increasing demand function w.r.t. discount $d_{i, t+n}$ by construction. The monotonicity is a standard assumption in econometrics (see e.g.,~\citep{Phillips2021}).
The final demand prediction at the future discount level $d_{i, t+n}$ is then the simple computation of~\eqref{eq:monotonic_demand_response}, given the parameterization of $\xi$ by $\phi_0$ and $\phi_1$. In order to provide demand predictions $q_{i,t+1:t+h}$ over the full forecast horizon $t+1:t+h$, the steps are repeated for all future discounts $d_{i,t+1:t+h}$ and the corresponding decoder output columns in $\kappa_{i,t+1:t+h}$.

\subsection{Near and Far Future Forecasts} \label{subsec:nearfar}
In our approach, we decompose the forecast problem into near future and far future.
The near future performance is of particular importance to our pricing use-case, hence we specialize our model architecture for this use case by 
having one decoder and one monotonic demand layer for each of the two forecast horizons. We set 5 week as the threshold for near future, and 5 to 20 week forecast will be considered as far future.

During training, we first train for seven epochs the decoder and monotonic demand layer for near future and freeze the 
decoder and demand layer for the far future. Then we freeze the components serving for near future and only train one final epoch 
for far future. We chose this specific training procedure to reduce training time, since the backpropagation of the error over the long forecast horizon turned out to be a significant cost factor. The near future loss only computes
for the first 5 weeks, i.e. the near future horizon; the far future loss is covering all the 20 weeks. Note that the model is used to predict for 26 weeks ahead using the far future decoder and demand layer, depending on extrapolating beyond the training horizon.

\begin{figure}
    \centering
    \includegraphics[width=\textwidth]{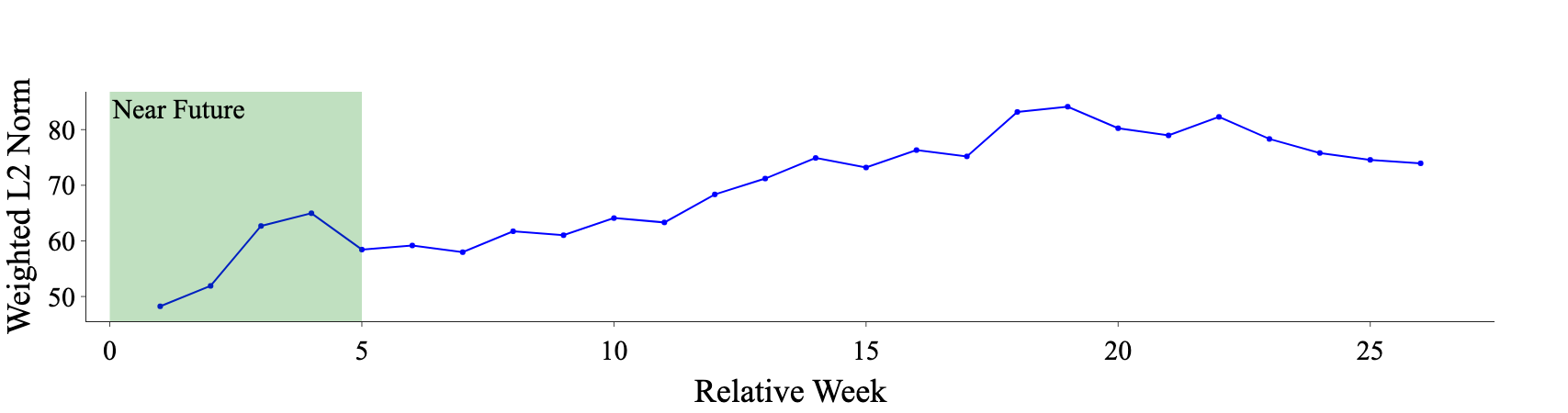}
    \caption{Demand model accuracy over a 26 weeks forecast horizon.}
    \label{fig:perf_over_time}
\end{figure}

In general, for forecasting tasks the further we forecast into the future the more deterioration there will be
in the forecast. From Figure~\ref{fig:perf_over_time}, we can see that the weighted $\ell_2$ norm, which we use 
in practice as an evaluation metric, is much smaller on average in the near future part compared to the far
future. The weighted $\ell_2$ norm
increases fast in the far future, which is expected. The far future is mainly used for long term effect
and to estimate trend, therefore the accuracy requirement for far future is not as high as the near future.

\subsection{Training}\label{sec:training}
In our training procedure, we follow mostly standard deep learning approaches. For the loss function that we use to train our model, we note that in our down-stream price optimization applicate, we are primarily interested in maximizing expected profit and profit is a (mostly) linear function. The linearity of the profit function allows us to work directly with the mean of~\eqref{eq:demand_model_prob}. Hence we use a loss function closely resembling the $\ell_2$-loss:

\begin{equation}\label{eq:loss}
\mathcal{L} = \sum_{j \in M} ( v( \hat{q}_{j}) - v(q_{j}))^2
\end{equation}
where $q$ is the demand and $\hat{q}$ the predicted demand and $v(x) = 1 + x + x^2/2 + x^3/6$ is the third-order Taylor approximation of $e^x$. This showed a stabilizing effect in the beginning of the training since the model became less sensitive to large errors. Note that we need to exponentiate, since we apply Box-Cox $\log$ transformation to our targets, see Table~\ref{tab:covariates}. Index $j$ ranges over all markets $M$.
The loss is not computed on the observed sales but on the inferred demand (see Section~\ref{sec:pecularities}). Recall from Section~\ref{subsec:nearfar} that we place a special emphasis on the near future, since pricing decisions are updated frequently and only prices for the next time unit will materialize. We train the encoder and short-term decoder jointly for most of the epochs. In the last epochs of training, we freeze the encoder and only train the decoder for predicting  the remaining far future weeks. This has the advantage to focus the encoder on the more important near-term forecasting problem but also has computational upsides as we do not need to train the high-dimensional decoder for the long-term. 

\subsection{Prediction}
The demand forecast model generates article level demand predictions for 14 different countries in the price range of $0$ to $70\%$ discount in 5 percentage point discount steps. The output of this large scale prediction run is a so-called \emph{demand grid} $\Upsilon \in R^{a \times c \times t \times d}$. The demand grid  $\Upsilon$ spans over $t = 26$ future weeks, up to $a = \num{1e6}$ articles, $c = 14$ countries and $d = 15$ discount levels. All predictions are generated via the global demand forecasting model introduced in Section~\ref{sec:model}. 
In order to produce article specific predictions, the model first loads the specific features of one article like demand  and price history as well as categorical features and other covariates~\ref{tab:covariates}. The model then predicts future demand for this specific article given the future discount level. Put differently, we answer the question: what would be the demand $q$ of an article $i$ in future week $t$ if we keep everything impacting demand constant but only change the price of the article by applying discount $d$. This makes it inherently a causal forecasting~\cite{bica2020estimating} problem as described in~\eqref{eq:demand_model_prob_causal}. 
The demand grid results in a large data batch with up to
\num{5.5e9} records and demand forecasts over the full range of future discount levels. Based on the discount dependent demand forecast grid, the down-stream price optimizer then selects an article specific future price trajectory that maximizes the achievable profit of the individual article in the bounds of certain business rules like stock levels and maximum discount steps. 

\section{Empirical Results}\label{sec:empirical_result}
The model is trained on the full Zalando article assortment, processing 4800 samples per batch. We choose 6 attention layers each in the encoder and decoder. Each multi head attention layer contains $h = 23$ attention heads. Including the heads, the dimension of the $Q, K$ and $V$ tensors is $h \times (j+1) \times d_k$ with 
$j = 51$\footnote{$j$ is the observational time horizon and selected to cover a complete year starting at index 0, considering weekly time resolution.} and $d_k = 4$\footnote{$d_k$ is a hyperparameter defining the dimension of $Q, K, V$.}.  For the monotonic demand layer, we choose seven segments of equal size 0.1. For the hyperparameter $\gamma$ in the loss~\eqref{eq:loss}, we choose the demand share of the associated market. In total, the model contains approximately \num{5.3e7} trainable parameters.

\begin{figure}
    \centering
\includegraphics[width=\textwidth]{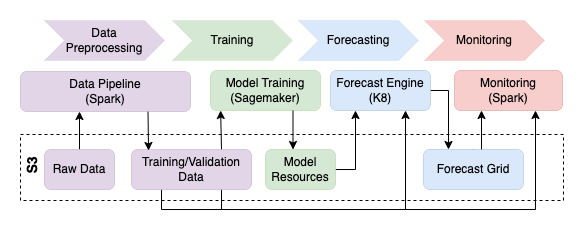}
    \caption{Forecasting Pipeline set-up in production.}
    \label{fig:pipeline}
\end{figure}

A standard time to train the model is about 45 hours on a four GPU instance (ml.g4dn.12xlarge) at a cost of approximately 275 US\$. We use a standard forecasting system set-up combining Spark and Amazon SageMaker~\citep{liberty20,wupper17,ZahariaXinEtAl16cacm}, depicted in Figure~\ref{fig:pipeline}. K8 refers to Kubernetes and S3 is the Amazon Web Service Storage Service.

\subsection{Accuracy metrics}\label{subsection:acc-metrics}
For measuring the accuracy of our demand forecast we need metrics that fulfill the following criteria:
\begin{enumerate}
    \item comparable between datasets with different number of articles
    \item weighted by how valuable an article is
    \item working well with our downstream optimization task.
\end{enumerate}
We define the \emph{demand error}
\begin{equation}\label{def:demand_error}
    \mathcal{D}_{T,h} = \sqrt{\frac{\sum_{i}\sum_{T=t+1}^{t+h}b_i(\hat{q}_{i,T} - q_{i,T})^2}{\sum_{i}\sum_{T=t+1}^{t+h}b_iq_{i,T}^2}}
\end{equation}
and the \emph{demand bias}
\begin{equation}\label{def:demand_bias}
    \mathcal{B}_{T,h} = \frac{\sum_{i}\sum_{T=t+1}^{t+h}b_i(\hat{q}_{i,T} - q_{i,T})}{\sum_{i}\sum_{T=t+1}^{t+h}b_iq_{i,T}}
\end{equation}
as the custom metrics to assess the accuracy of our forecasting problem. Here, $t$ is the last timepoint in the training set, $h$ denotes the forecast horizon, 
$\hat{q}_{i,T}$ is the prediction for article $i$ at timepoint $T$, $q_{i,T}$ is the corresponding true demand and $b_i$ is
the black price of article $i$.

The demand error $\mathcal{D}_{T,h}$ measures how large our forecasting error is on a single article level. We weight this by black price
$b_i$ to place a higher importance on more expensive articles as a proxy for the overall value for Zalando. As these typically generate more profit per sale, an error
in the demand forecast for these articles is more costly for us. Scaling ensures that even in our setting, where the set of forecasted articles  and sold changes from week to week, we obtain comparable values. In contrast, the demand bias indicates whether we are on average over or
underpredicting while using the same weighting and scaling as the demand error.

We also experimented with (weighted and scaled) versions of more common accuracy metrics, like MAPE and MAE, and ran simulations how this changes affect the downstream optimization task.
It turns out that outliers are an important factor for the performance so that a metric that resembles the RMSE more closely will have the best result with respect to 
the downstream pricing application.
This feature is also the biggest drawback of our demand error metric as we observe that the error is largely driven by a small portion of the data. Especially
forecast accuracy on articles with low and sparse demand (which is the majority of our articles) is not adequatly adressed by these metrics.

\subsection{Model benchmarking}
To illustrate accuracy of our transformer model we present a part of the empirical results we gathered. In particular we compare the model described above with a naive forecast that simply repeats the last observation, a tree-based model~\citep{forecastingWTrees} using LightGBM~\citep{lightGBM} and the autoregressive recurrent neural network DeepAR~\citep{salinas2020deepar} as implemented in GluonTS~\citep{alexandrov2019gluonts} for which we used the mean forecast as a point forecast. For the experiments, we chose 10 weeks in 2022, using all sales data from Germany from the category Textil. In Zalando this is the largest category, and performance here is highly indicative of the performance on the whole assortment. We focus primarily on the performance of the first weeks forecast. Table \ref{table:emp-results} shows the average metrics in the given weeks, along our own metrics defined as in Section~\ref{subsection:acc-metrics}, we also calculated more widely used metrics, namely RMSE and MAPE, where, to deal with zeros, we added 10 to both predicted and true demand values.

\begin{table}
\centering

\begin{tabular}{l cc   cc  cc  cc}
 
    \toprule
Forecaster & \multicolumn{2}{c}{Demand Error} & \multicolumn{2}{c}{Demand Bias} & \multicolumn{2}{c}{RMSE} & \multicolumn{2}{c}{MAPE} \\
\midrule
& Mean & Std & Mean & Std & Mean & Std & Mean & Std\\
Naive Forecaster & 0.603 & 0.135 & 0.079 & 0.161 & 11.754 & 4.038 & 0.088 & 0.011\\
lightgbm & 0.489 & 0.066 & 0.296 & 0.127 & 9.737 & 2.941 & 0.142 & 0.005 \\
DeepAR & 0.593 & 0.126 & 0.090 & 0.156 & 12.443 & 4.224 & 0.092 &  0.011 \\
Transformer & \textbf{0.449} & 0.04 &  \textbf{- 0.047} & 0.064 & \textbf{9.588} & 2.725 & \textbf{0.081} & 0.007\\
\bottomrule
\end{tabular}
\caption{Model Performance}\label{table:emp-results}
\caption*{Comparison of our transformer models with commonly used alternative approaches and a naive baseline. The metrics are calculated over 10 different weeks in 2022 and measure the performance of the predictions for the first week. We report here the mean and standard deviation over the 10 weeks.}
\end{table}

We observe that our transformer model performs best in terms of demand error and RMSE. While tree based models seem to be a close competitor in terms of demand error and were much faster to train, we saw a significantly higher bias. In terms of demand bias and MAPE, both Deep Learning models are very close to each other, but for RMSE again lightgbm and our transformer model have a clear advantage. It is worth mentioning that the naive baseline performs quite competitively on the one week forecast. A more detailed analysis reveals that especially for high selling articles using the naive forecast poses a strong baseline. Obviously during periods of high variation, as they happen at the beginning and the end of sales events, the naive forecast becomes almost useless. During the end of season sale in 2022 for example we observed a demand error of $0.877$ for the naive forecaster while our transformer model reached a much more stable demand error of $0.491$.

\subsection{On the Benefits of Transformer-based Forecasting: First Results}

In the overall literature on forecasting, conclusive evidence of outsized gains in predictive accuracy of transformer-based models is not yet available. On the one hand, results on publicly available data sets are positive~\citep{lim2021temporal}, but not as overwhelmingly positive as in natural language processing (NLP) where transformers dominate. Careful comparisons with strong baselines (e.g.,~\citep{oliver22,tileForecaster}) place additional scrutiny on these results. We remark that transformer-based forecasting approaches tend to distinguish themselves by side-aspects such as long-term forecasting or interpretability whereas in NLP, the improvements in the core task are outsized. 

On the other hand, on closed data sets, convincing evidence exists for the superiority of transformer-based forecasting models, e.g.,~\citep{carson20,dhruv2}. Our results above point into a similar direction. So, a natural question to ask is whether (and how) the success of transformer-based models can be associated to the nature of these closed source data sets and in particular the abundace of data available for companies like Zalando. To make progress in answering this question, we considered an often-cited reason for the success of transformers, so-called scaling laws (e.g., \cite{ScalingLaws, FewShotLearners}). These scaling laws state that the predictive performance on a test set continues to increase for transformers along various dimension including, for example, the size of the training set.

\begin{figure} 
\includegraphics[width=\textwidth]{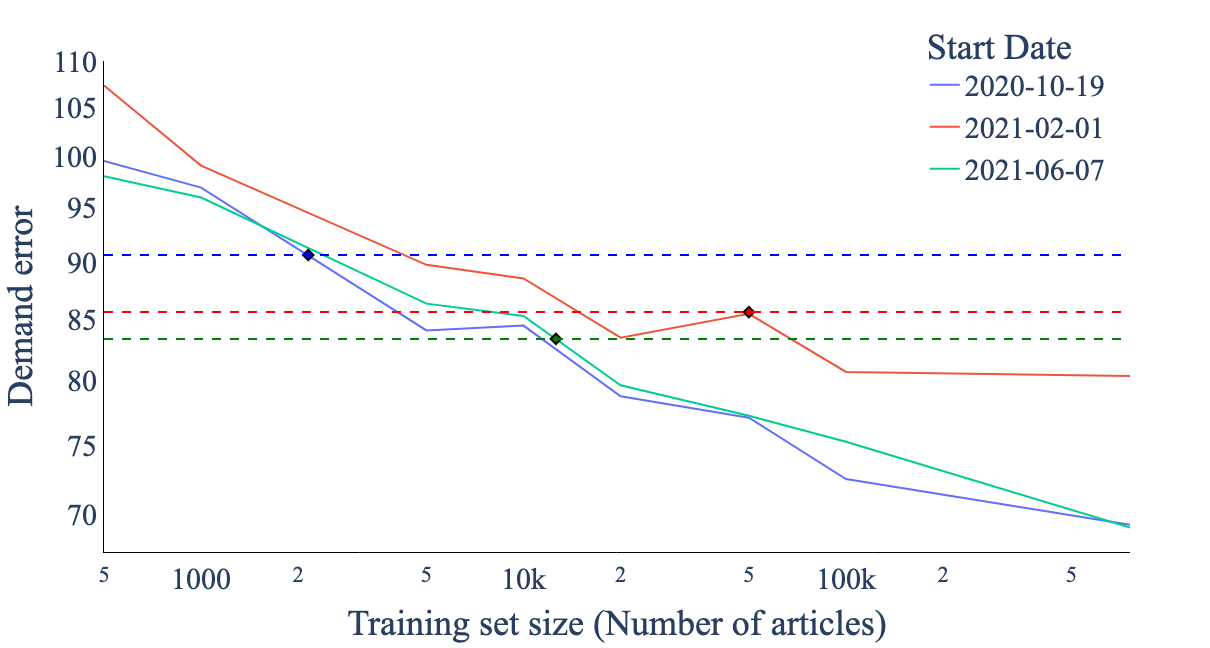} 
\caption{Log-log plot of the training-set-size-to-demand-error relationship for three different start dates with constant test set. The dashed lines represent the forecast performance of the naive forecaster (last observation in training horizon).} 
\label{fig-scaling-law} 
\end{figure}

Concretely, we attempt to replicate scaling laws for transformers in forecasting, e.g., in \citep[Figure 1]{ScalingLaws} and whether we can observe level effects (see Figure \ref{fig-scaling-law}).

\paragraph{\textbf{Experimental Setup}}
\begin{itemize}
    \item We chose three different forecast start dates distributed over 2021. 
    \item We use 208 weeks of history to forecast 5 weeks for each each start date.
    \item On each of the three start dates, we use a representative subset of roughly 760K different articles and always use these as the reference point for testing with regards to the Demand Error. In this sense, we will have two layers of generalization present: The first stemming from predicting future demand from time series whose history is present in the training data and the second corresponding to completely unseen parts of the assortment. 
    \item For smaller training sets we ensure that we always have a comparable number of batches per epoch (and hence, comparable training time) by iterating over the same time series several times.
    \item All other hyperparameters of the training procedure (batch size, cloud instance, learning parameters,...) remain constant and mimic our production setup.
    \item We included a naive forecast based on the last week's observation in the training horizon to check under which train dataset sizes our model starts to outperform. 
\end{itemize}

The demand error for each train dataset size (in terms of the number of time-series) as well as different start dates are shown in Fig~\ref{fig-scaling-law}. 
The naive forecast exhibits the expected behaviour and its performance is naturally constant in the size of the training set. Interestingly, the naive forecaster is already outperformed by training our model on a small fraction of the assortment (between 0.2\% and 6.6\% for the three start dates under consideration). For two out of the three start dates, the scaling law holds well. Similar scaling laws on public data sets in forecasting with other transformer-based models have not been reported so far (and rather the contrary on publicly available data sets~\citep[Figure 2]{oliver22}).

Clearly, the above experiment is only a first step in understanding a potential superiority of the transformer-based approach. Our data sets and associated training times are still humble compared to what state-of-the-art NLP-models use, so using more available data (e.g., by going further back in time) presents an obvious opportunity. We believe that this is the first time that evidence for scaling laws for transformers in forecasting was presented.

\section{Related Work}\label{sec:rel_work}
Our demand model is a customized transformer-based forecasting model \cite{vaswani2017attention}. When we launched it in 2019, the current rich landscape of transformer based forecasting models was not yet available. So, we drew inspiration from advances in deep-learning architectures applied to sequential learning problems \cite{sutskever2014sequence} and from the success of the transformer architecture in the field of natural language processing \cite{devlin2018bert}. In this, our development is similar to other online retailers~\citep{carson20} who worked on transformer-based time series models concurrently to our work, but prior to the publicly available state of the art. For the latter, remarkable current transformer models include~\cite{lim2021temporal,NEURIPS2019_6775a063,informer20}. But, it remains an open question how well these fare against proven state of the art deep-learning based models (e.g.,~\citep{salinas2020deepar,oreshkin2019n}): We have not yet compared ourselves against these baselines for practical reasons and the introduction of state of the art open source libraries~\citep{alexandrov2019gluonts} will facilitate this. We refer to~\citep{benidis22} for an overall overview on deep-learning based forecasting models.


Note that the approach described here is a point forecast. While clearly not best practice, the (piece-wise) linearity of the profit function used down-stream~\citep{Falcon20} partially remediates this methodological short-coming. We point to~\citet{adele21} how to turn our model into a flexibly parameterized probabilistic forecasting model. Other generic approaches amend themselves readily to our model such as conformal forecasting~\citep{conformal} or quantile-based approaches~\citep{pmlr-v89-gasthaus19a,pmlr-v151-kan22a}. We note further that we could handle the intermittency of the demand data more directly, e.g.,~\citep{renewal,JEON20221415}. This is future work. 

Modelling the demand response function as a piece-wise linear function is, to the best of our knowledge, novel. However, we remark that in other contexts, this has been proposed before in forecasting~\citep{pmlr-v89-gasthaus19a} and the general theme of modelling the monotonic relationship between price and demand is explored recently e.g.,~\citep{asos22} although not for deep-learning based models.

Finally, we remark that our model~\eqref{eq:demand_model_prob} is an over simplification in the sense that we ignore cross-item relationships or the hierarchical relationship of our data. Recent work for multi-variate forecasting in a deep learning context (e.g.,~\citep{salinas2019high,bezene20,rasul2021multivariate}) and hierarchical forecasting (e.g.,~\citep{rangapuram21a,Theodosiou:2021a,sharq}) exists, but we leave the application of it for future work.


\section{Practical Considerations}\label{sec:practice}
We have so far described the data and the model assuming a mostly static setting. In reality however, forecasts need to be made available on a weekly schedule in an ever-changing environment. This leads to additional challenges which in turn requires a number of additional artefacts. We summarize this briefly on a high-level.

\begin{figure}
\begin{center}
\includegraphics[width=\textwidth]{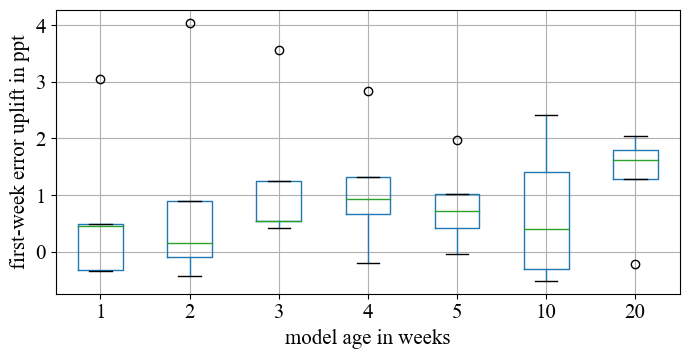}
\caption{Effect of retraining on accuracy: The plot shows how much accuracy the model loses if not retrained regularly, provided a certain level of randomness introduced by the training process.}
\label{fig:retraining}
\end{center}
\end{figure}

First, we must be able to judge whether our live forecasting accuracy is in line with the accuracy we have in our backtests. For this, we have extensive monitoring metrics that compare incoming live sales data with our forecasts and alarm us, if we observe a deterioration in practice. Second, we must be able to roll-out model changes (e.g., adjustments of hyper-parameters, additional covariates or library updates would be small examples). For this, a deployment pipeline is used with a large suite of test scenarios that cover aspects such as accuracy or training \& inference speed in a variety of scenarios. Third, we monitor the input data distribution to scan it for distribution shifts. Note that we could use this for example for sophisticated re-training scheduling schemes, however, experiments reported in Figure~\ref{fig:retraining} have shown that a weekly retraining scheme offers best results. 

Finally, we note that the demand forecasting model described here, is only one input of more into a discount optimizer~\citep{Falcon20}. While our demand forecasts and price elasticities have some value in themselves, the most important artefact are discount decisions. We are only at the start of the journey to understand how forecast accuracy changes attribute to price changes in the complicated interplay between multiple models and (price) decisions that change the environment (and for which backtesting is hence hard). This is particularly involved because our systems evolves dynamically over the course of the seasonal cycle.


\section{Conclusion}\label{sec:concl}
We have presented the demand forecasting model employed by Zalando SE, one of Europe's leading fashion online retailer. This global, transformer-based model delivers demand forecasts week in and week out since its initial inception in 2019, thereby predating some of the existing literature on transformer-based forecasting models by at least a year. We showed that this model is competitive against other methods. We also provide a first explanation of the superiority of transformer-based models by testing for scaling laws, as our closed data set contains millions of time series.

The incorporation of price or discount is the main modelling innovation that we presented here and it also presents the largest opportunity for future work. We are in fact most interested in causal or counterfactual forecasts -- because these allow the most meaningful downstream decisions. Work on causal forecasting exists (e.g.,~\citep{causalforecasting,causalforecasting2}), but is in its infancy. We believe that much further work is needed and hope that the exposition of our problem presented here will help to inspire such work further.

\section*{Acknowledgments} This article represents the effort of a larger group with everyone contributing to the manuscript and methodology in unique and impactful ways. We are grateful for their contributions without which we would not have this manuscript (and neither the model running in production at Zalando). We want to especially thank Tofigh Naghibi, who inspired the team to apply neural networks to our time series forecasting problem back in 2017. He did not only provide the initial idea but also contributed the first working prototype of the transformer based forecasting model. His work heavily influenced the team and was setting the foundation for a successful deep learning based forecasting application.

\bibliographystyle{plainnat}
\bibliography{references}

\end{document}